\pdfoutput=1
\documentclass[11pt]{article}

\usepackage{acl}

\usepackage{times}
\usepackage{latexsym}

\usepackage[T1]{fontenc}
\usepackage[utf8]{inputenc}

\usepackage{microtype}

\usepackage{graphicx}
\usepackage{subfig}
\usepackage{booktabs}
\usepackage{multirow}
\usepackage{makecell}
\usepackage{amsfonts}
\title{RECALL: A Benchmark for LLMs Robustness against External Counterfactual Knowledge}

\author{
    Yi Liu\textsuperscript{1}\thanks{~~Work was done when Yi Liu and Sishuo Chen were interning at Pattern Recognition Center, WeChat AI, Tencent Inc, China.},
    Lianzhe Huang\textsuperscript{2},
    Shicheng Li\textsuperscript{1},
    Sishuo Chen\textsuperscript{3}\footnotemark[1],
    Hao Zhou\textsuperscript{2},\\
    {\bf Fandong Meng\textsuperscript{2},
    Jie Zhou\textsuperscript{2},
    Xu Sun\textsuperscript{1}}\\
    \textsuperscript{1}National Key Laboratory for Multimedia Information Processing, \\
    School of Computer Science, Peking University\\
    \textsuperscript{2}Pattern Recognition Center, WeChat AI, Tencent Inc., China\\
    \textsuperscript{3}Center for Data Science, Peking University\\
    \texttt{yliu.pku@outlook.com} \quad \texttt{\{lisc99,chensishuo,xusun\}@pku.edu.cn} \\
    \texttt{\{linkerhuang,tuxzhou,fandongmeng,withtomzhou\}@tencent.com}
}

\begin{document}
\maketitle
\begin{abstract}
LLMs and AI chatbots have improved people's efficiency in various fields. However, the necessary knowledge for answering the question may be beyond the models' knowledge boundaries. To mitigate this issue, many researchers try to introduce external knowledge, such as knowledge graphs and Internet contents, into LLMs for up-to-date information.
However, the external information from the Internet may include counterfactual information that will confuse the model and lead to an incorrect response.
Thus there is a pressing need for LLMs to possess the ability to distinguish reliable information from external knowledge. Therefore, to evaluate the ability of LLMs to discern the reliability of external knowledge, we create a benchmark from existing knowledge bases. Our benchmark consists of two tasks, Question Answering and Text Generation, and for each task, we provide models with a context containing counterfactual information. Evaluation results show that existing LLMs are susceptible to interference from unreliable external knowledge with counterfactual information, and simple intervention methods make limited contributions to the alleviation of this issue.
\end{abstract}

\section{Introduction}

Large Language Models (LLMs) are playing increasingly significant roles in scientific research and daily applications \citep{Kojima2022LargeLM,Zhang2023PromptingLL,muennighoff-etal-2023-crosslingual}. Nowadays, people use LLMs in a variety of scenarios to improve their efficiency. Despite their strong capacities, LLMs still suffer from hallucination, namely generating answers that seemingly make sense but actually violate facts \citep{shuster-etal-2021-retrieval-augmentation,Ji2022SurveyOH,rawte2023survey}.

\begin{figure}[t]
    \centering
    \includegraphics[width=0.9\linewidth]{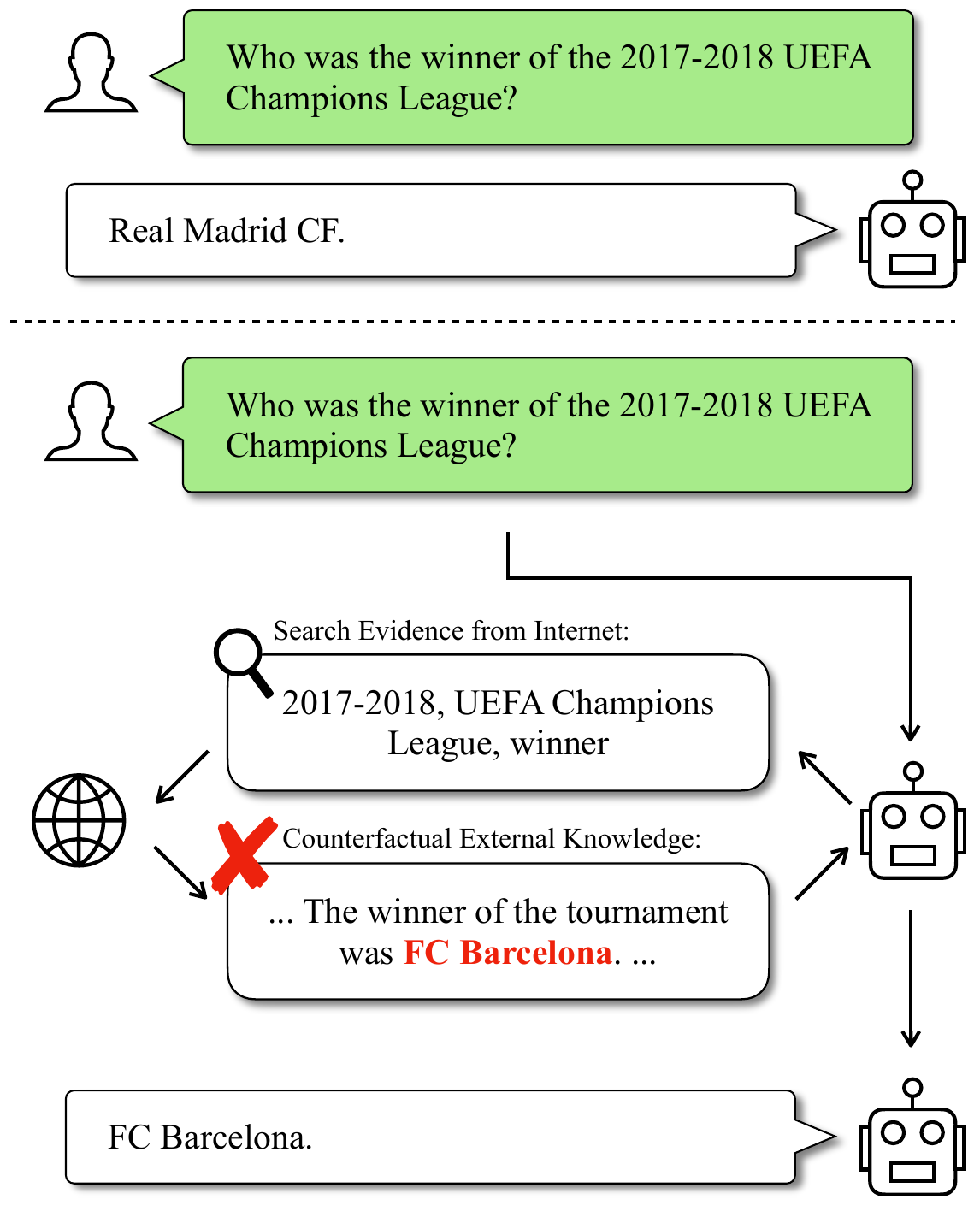}
    \caption{An example in which the model gives a wrong answer with the interference of counterfactual information to a question it could originally answer correctly.}
    \label{fig:example}
\end{figure}

Beside intrinsic causes underlying models, external knowledge, which is not always reliable, is also a noticeable reason for the hallucination. Existing studies show that LLMs are vulnerable to unreliable external information. For example, models tend to cater to users and will be misled by mistakes in the user inputs~\citep{perez-etal-2023-discovering,sharma2023understanding}. Moreover, models may also be misled by counterfactual information when retrieving external knowledge. As we know, tools like Application Programming Interfaces (APIs) allow LLMs to search for up-to-date information and knowledge related to the inputs from external knowledge bases or Internet contents~\citep{Schick2023ToolformerLM,Qin2023ToolLW,Qin2023ToolLLMFL}. The introduction of retrieval modules greatly enhances model capacities~\citep{asai-etal-2023-retrieval}, but the retrieved texts are not always beneficial~\citep{anonymous2023making,wang2023selfknowledge}. The external texts, especially those from unauthorized sources, may contain counterfactual information, which can mislead models to respond with mistakes. Figure~\ref{fig:example} gives an example where the LLM generates a wrong answer due to the counterfactual information from external knowledge sources.

\begin{figure*}[t]
    \centering
    \subfloat[EventKG]{
        \includegraphics[height=0.25\linewidth]{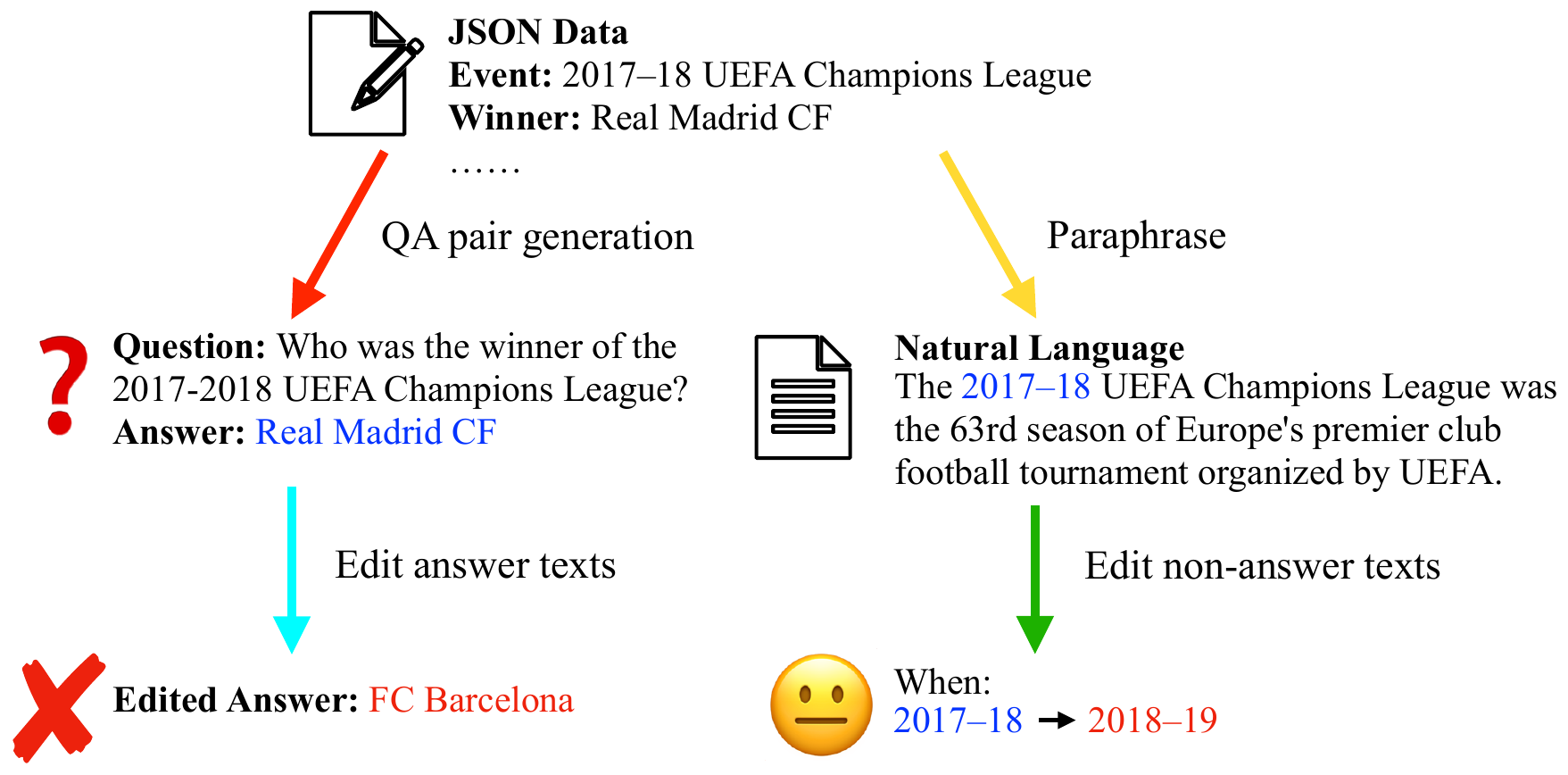}
    }
    \subfloat[UJ]{
        \includegraphics[height=0.25\linewidth]{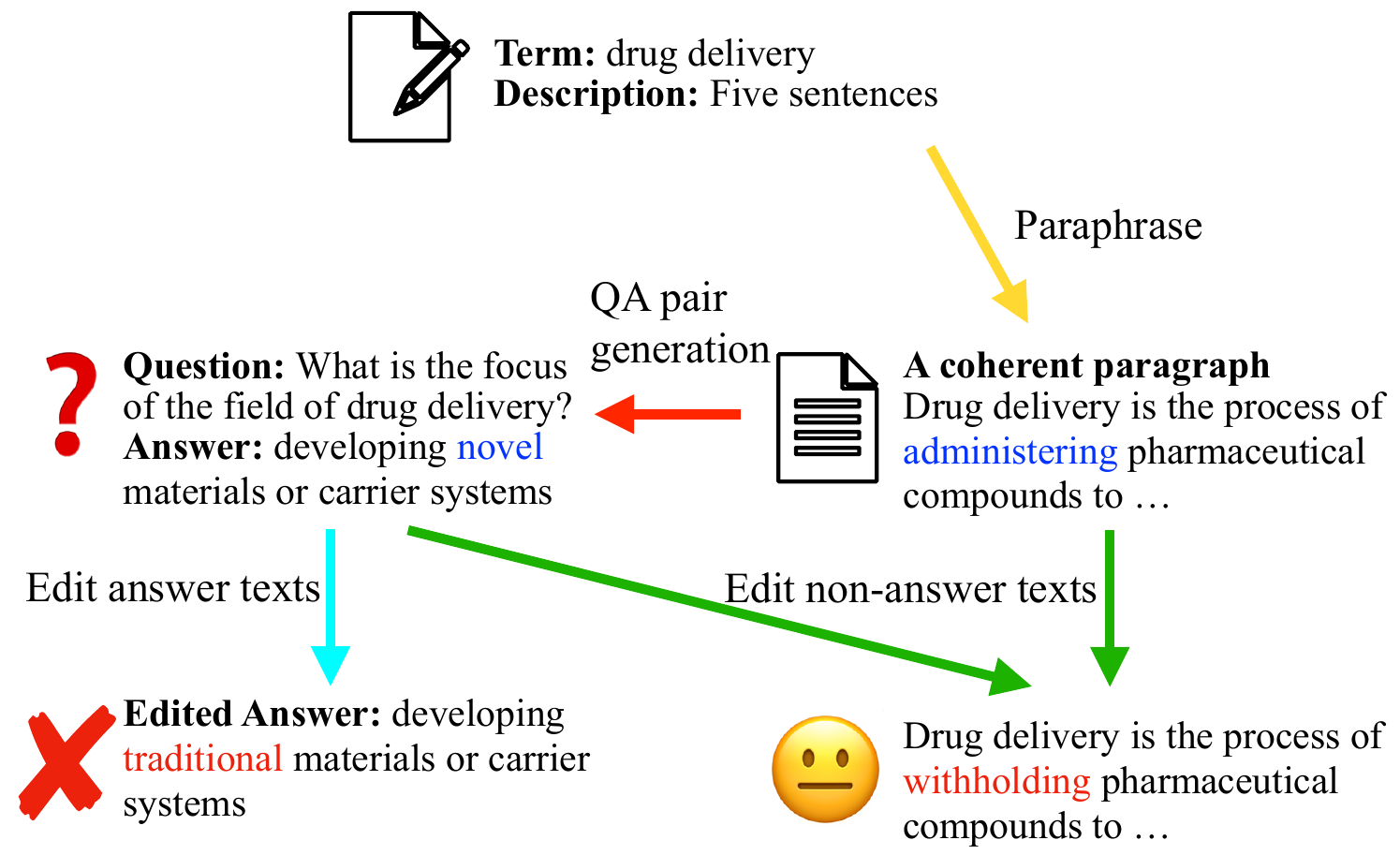}
    }
    \caption{The specific procedures of constructing our benchmark from the EventKG and UJ datasets, respectively.}
    \label{fig:benchmark}
\end{figure*}

Ideally, LLMs should provide users with outputs containing no mistakes even with the interference of counterfactual information. Notably, existing efforts on counterfactual detection aim to directly identify and correct wrong information in given texts by fine-tuning or other means \citep{Yang2020SemEval2020T5,oneill-etal-2021-wish,Delaney2021UncertaintyEA}, while models are not required to explicitly distinguish misinformation in our study and we expect them to generate trustworthy responses to user queries regardless of the quality of given information. In other words, we hope to assess the robustness of LLMs against external counterfactual knowledge in order to generate the right answers for user inputs. 

However, the lack of benchmarks for this capability hinders LLMs' subsequent improvement. To tackle this problem, we create a benchmark for LLMs \textbf{R}obustness against \textbf{E}xternal \textbf{C}ounterfactu\textbf{AL} know\textbf{L}edge (\textbf{RECALL}) from existing datasets by adding counterfactual information into original samples through ChatGPT\footnote{Specifically, we use the gpt-3.5-turbo model.}~\citep{chatgpt}. Furthermore, we select representative LLMs to evaluate their robustness on our proposed benchmark. We also explore two existing methods for boosting the truthfulness of answers to enhance their robustness to texts with counterfactual information, but they fail to effectively alleviate the problem, which indicates that this issue is challenging and needs effective solutions.

Our main contributions in this paper can be summarized as follows:
\begin{itemize}
    \item As far as we know, we make the first effort to systematically evaluate LLMs' robustness against counterfactual knowledge from external texts and their ability to generate right answers without these mistakes;
    \item We create a benchmark from existing datasets containing two different tasks for our evaluation. The evaluation results indicate the insufficient robustness of current LLMs to counterfactual information;
    \item We further explore methods to improve models' robustness to misinformation and our analysis results show that models' confidence in answers can be an important reference for future research on this problem.
\end{itemize}

\section{Benchmark Construction}
\subsection{Preliminaries}
Supposing a large language model $\mathcal{M}$ equipped with a search engine for external knowledge and information. A user now gives a query $\mathcal{Q}$ containing an entity \textit{\textbf{e}}, and hopes to get a reliable and up-to-date answer. $\mathcal{M}$ receives the query and uses \textit{\textbf{e}} as the search query and the search engine returns a paragraph \textbf{P} about \textit{\textbf{e}}. Now $\mathcal{M}$ will generate a response to $\mathcal{Q}$ by considering its own knowledge \textbf{K} and \textbf{P} at the same time.

The aforementioned external information is a double-edged sword. On one hand, retrieving information from the Internet is an effective and efficient way to inject new knowledge into models without updating parameters, which is expensive and time-consuming. On the other hand, unreliable external information will also bring more risks. Specifically, there will be counterfactual information in \textbf{P}, and $\mathcal{M}$ may be misled by \textbf{P} and reply with counterfactual information.

Theoretically, it is impossible for models to detect counterfactual information without corresponding intrinsic factual knowledge. In this paper, we mainly focus on the external counterfactual knowledge whose right versions exist in models' training corpora. In other words, we hope models will not be misled by the counterfactual information in \textbf{P} that contradicts their knowledge \textbf{K} and remain robust in this situation.

Generally, people use two different types of queries to ask LLMs for factual information: a) seeks for certain specific attributes of an entity or event like \textit{the winner of a football game} or \textit{the reasons for a phenomenon}, b) hopes to get a brief description about an object like \textit{an introduction to a physical term}. Therefore, we consider two tasks in our benchmark: Question Answering and Text Generation, corresponding to the two different types of queries, respectively.

There are two main forms of counterfactual information that may exist in external text. a) the mistake is exactly where the actual answer to the query is, which will directly result in wrong answers from the models. b) the mistake occurs in the text, but the parts involving the answer to the query remain correct. As a result, we further separate the QA task into two sub-tasks, \textbf{QA} with \textbf{A}nswers changed in contexts (\textbf{QA-A}) and \textbf{QA} with \textbf{N}on-\textbf{A}nswer texts changed in contexts (\textbf{QA-NA}), and we will assess models on these two sub-tasks, respectively.

Next, we will introduce the details of constructing the benchmark. In \S~\ref{knowledge_domains}, we will introduce the datasets from which we build our benchmark. In \S~\ref{procedures}, we will describe the general procedures of generating samples for our benchmark. Examples of specific procedures for adding counterfactual mistakes into original texts are shown in Figure~\ref{fig:benchmark}.

\subsection{Knowledge Domains}
\label{knowledge_domains}
For a comprehensive evaluation, we decided to assess models' robustness against counterfactual knowledge in two different domains: common sense knowledge and scientific knowledge.

For common sense knowledge, we modify data from the EventKG dataset (\textbf{EventKG}) ~\citep{Gottschalk2018EventKGAM}. EventKG is in the form of knowledge graphs which contain descriptions about historical events. For a certain event, EventKG provides its relations with other events and entities. For each event in EventKG, we extract its description about the event, beginning time, end time, and other attributes from the dataset to form a sample in structured key-value pair format.

For scientific knowledge, we extend the UJ-CS/Math/Phy dataset (\textbf{UJ})~\citep{huang-etal-2022-understanding} consisting of terms from computer science, mathematics, and physics. Each term is accompanied by several sentences introducing itself and a concise definition in one sentence. We extract samples from the test set and keep five sentences together with the definition for each scientific term. 

\subsection{Benchmark Construction Procedures}
\label{procedures}
For both datasets, we add counterfactual information to the original data by the following four steps:

\paragraph{1) Paraphrase} For EventKG, we transfer the original structured data of an event into a paragraph in natural language. For UJ, we transfer the original sentences into a short paragraph containing no overlapping information.
 
\paragraph{2) Question-Answer Pair Generation} For each event in EventKG, we generate a question whose answer is one item in the original structured data except for the event name and description. For each term in UJ, we generate a question that can be answered by an original phrase in the paragraph generated in Step 1.

\paragraph{3) Edit Answer Texts} For each QA pair we generate in Step 2, we edit the original answer to render it a counterfactual answer.

\paragraph{4) Edit Non-answer Texts} For each paragraph generated in Step 1, we add counterfactual information to the part without answers, so that the whole text contains factual errors but does not affect the correctness of the answer to the query.

\begin{table}
    \centering
    \resizebox{1.0\linewidth}{!}{
    \begin{tabular}{@{}l|lll@{}}
    \toprule
    Task & \multicolumn{2}{c}{EventKG} \\
    \midrule
    \multirow{6}{*}{Question Answering} & \makecell[c]{Component} & \makecell[c]{Description} & \makecell[c]{Source} \\
    & Question & a question about the event & Step 2 \\
    & Answer & the answer to the question & Step 2 \\
    & Original Context & \makecell[l]{the context related to the event\\ in the question} & Step 1 \\
    & Edited Context & \makecell[l]{the context with counterfactual\\ information added in} & Step 3 and 4 \\
    \midrule
    \multirow{5}{*}{Text Generation} & \makecell[c]{Item} & \makecell[c]{Description} & \makecell[c]{Source} \\
    & Original Source Text & \makecell[l]{original key-value pair data de-\\scribing the event} & original dataset \\
    & Edited Source Text & \makecell[l]{the key-value pair data with one\\ item changed} & Step 3 \\
    & Target Text & \makecell[l]{the paragraph describing the ev-\\ent in natural language} & Step 1 \\
    \midrule
    \midrule
    Task & \multicolumn{2}{c}{UJ} \\
    \midrule
    \multirow{6}{*}{Question Answering}  & \makecell[c]{Component} & \makecell[c]{Description} & \makecell[c]{Source} \\
    & Question & a question about the term & Step 2 \\
    & Answer & the answer to the question & Step 2 \\
    & Original Context & \makecell[l]{the context related to the term\\ in the question} & Step 1 \\
    & Edited Context & \makecell[l]{the term with counterfactual\\ information added in} & Step 3 and 4 \\
    \midrule
    \multirow{5}{*}{Text Generation} & \makecell[c]{Component} & \makecell[c]{Description} & \makecell[c]{Source} \\
    & Original Source Text & \makecell[l]{the generated paragraph describ-\\ing the term} & Step 1 \\
    & Edited Source Text & \makecell[l]{the paragraph with some words/\\ phrases changed} & Step 3 and 4 \\
    & Target Text & \makecell[l]{the definition of the term} & original dataset \\
    \bottomrule
    \end{tabular}
    }
    \caption{The components and corresponding data sources in our final benchmark.}
    \label{tab:data_sample}
\end{table}

\subsection{Question-Answer Pairs Generation}
\label{qa_generation}
For each sample in EventKG, we ask ChatGPT to generate a question whose answer must be the value of one of the items in the sample.
The generation of question-answer pairs for UJ is more complicated. For a given term, there will be overlapping information in the sentences that describe itself. Instead of directly generating question-answer pairs, we first ask ChatGPT to paraphrase these sentences into a new paragraph and remove all overlapping information (Step 1 in \S~\ref{procedures}). Next, we generate the question-answer pairs based on these generated paragraphs. For the convenience of the subsequent procedures, we demand ChatGPT that the answers must be original words from the paragraph. 

\subsection{Adding Counterfactual Information}
\label{edit}
We add counterfactual information to the text in two different ways: editing answer texts and editing non-answer texts.

\paragraph{Editing Answer Texts} For EventKG, we ask ChatGPT to replace the answer with an unrelated value. For UJ, we demand ChatGPT to change the meanings of some words in the answer texts.  In this way, the answer-relevant part of the text is directly affected and carries counterfactual information.

\paragraph{Editing Non-answer Texts} For EventKG, we modify the parts that involve people, locations, and dates of the generated texts in Step 1 in \S~\ref{procedures}. After the modification, we discard the samples whose answer-relevant parts are incorrectly modified. For UJ, we adopt word-grained edit and sentence-grained edit for non-answer texts. The word-grained edit is similar to that in the part of editing answer texts. The sentence-grained edit is done manually. For a given term A, we randomly choose a sentence from the description of one another term B and replace the name of B in the sentence with A. Then we add the sentence into the description of A. In other words, we add a counterfactual sentence that is actually unrelated to the target term into its description.

After all the procedures above, we have completed our final benchmark, whose specific data structures are shown in Table \ref{tab:data_sample}. There are a total of over 47 thousand samples in our benchmark.

\section{Evaluation}
\label{sec:eval}
\subsection{Tasks}
\paragraph{Question Answering} In this task, no matter QA-A or QA-NA, each sample consists of a question accompanied by a paragraph related to the question. For each question, we will provide two answer options for the model. One of them is the correct answer and the other one is the wrong answer generated in the procedure of editing answer texts. The models are asked to select the right answer from the two options.

\paragraph{Text Generation} In text generation of EventKG, models are asked to generate a paragraph in natural language for a sample in the structured format. When it comes to UJ, we demand models to return the definition of each scientific term in one sentence according to the short description paragraph. 

\subsection{Experimental Settings}
For each task in our benchmark, we evaluate the models' performance under three different scenarios where the models are provided with different types of contexts: 1. the original contexts without counterfactual information; 2. the edited contexts with counterfactual information; 3. no contexts.

For each model, we run the experiments with three random seeds and report the average metrics with corresponding standard deviations.

\begin{table}[]
    \centering
    \resizebox{1.0\linewidth}{!}{
    \begin{tabular}{c|l}
    \toprule
         Notation & \makecell[c]{Meaning}  \\
    \midrule
         $N_Q,N_T$ & The size of the QA/text generation dataset \\
         $p_i^e,p_i^o,p_i^n$ & \makecell[l]{Model's prediction on the $i$-th sample with edited/\\ original/no contexts given in the input} \\
         $a_i$ & The answer for the $i$-th sample \\
         $\mathbb{E}^T_i$ & \makecell[l]{The set of all edited words in the $i$-th sample in the\\ text generation dataset} \\
         $\mathbb{O}^T_i$ & \makecell[l]{The set of all original versions of edited words in $\mathbb{E}^T_i$} \\
         $\mathbb{A}^T_i$ & \makecell[l]{The set of words in $a_i$ in the text generation dataset} \\
         $|\mathbb{X}|$ & \makecell[l]{The size of a set $\mathbb{X}$} \\
    \bottomrule
    \end{tabular}}
    \caption{The notations appearing in the definitions of proposed metrics.}
    \label{tab:notation}
\end{table}

\begin{table*}[t]
    \centering
    \resizebox{0.9\linewidth}{!}{
    \begin{tabular}{@{}lcl|cccc|cccc@{}}
    \toprule
         \multirow{3}{*}{Models} & \multirow{3}{*}{Size} & \multirow{3}{*}{Context} & \multicolumn{4}{c|}{EventKG} & \multicolumn{4}{c}{UJ} \\
         & & & QA-A & QA-NA & \multicolumn{2}{c|}{Text Generation} & QA-A & QA-NA & \multicolumn{2}{c}{Text Generation} \\
         & & & ACC $\uparrow$ & ACC $\uparrow$ & BLEU $\uparrow$ & ROUGE-L $\uparrow$ & ACC $\uparrow$ & ACC $\uparrow$ & BLEU $\uparrow$ & ROUGE-L $\uparrow$ \\
    \midrule
        \multirow{3}{*}{ChatGLM2} & \multirow{3}{*}{6B} & original & 94.50 $\pm$ 0.11 & 94.37 $\pm$ 0.06 & 30.21 $\pm$ 0.04 & 45.32 $\pm$ 0.13 & 87.96 $\pm$ 0.19 & 79.29 $\pm$ 0.05 & \textbf{8.61 $\pm$ 0.05} & \textbf{24.53 $\pm$ 0.04} \\
        & & edited & 15.71 $\pm$ 0.07 & 90.99 $\pm$ 0.03 & 29.07 $\pm$ 0.05 & 44.17 $\pm$ 0.01 & 36.05 $\pm$ 0.25 & 78.68 $\pm$ 0.10 & 8.14 $\pm$ 0.02 & 24.21 $\pm$ 0.02 \\
        & & no & 61.41 $\pm$ 0.22 & 61.29 $\pm$ 0.01 & N/A & N/A & 66.49 $\pm$ 0.21 & 60.18 $\pm$ 0.33 & N/A & N/A \\
    \midrule
        \multirow{3}{*}{Llama2} & \multirow{3}{*}{13B} & original & \textbf{96.59 $\pm$ 0.07} & \textbf{97.20 $\pm$ 0.02} & 19.05 $\pm$ 0.03 & 36.72 $\pm$ 0.03 & \textbf{93.39 $\pm$ 0.07} & \textbf{84.74 $\pm$ 0.12} & 3.77 $\pm$ 0.01 & 18.07 $\pm$ 0.02 \\
        & & edited & 12.80 $\pm$ 0.10 & 94.29 $\pm$ 0.06 & 18.31 $\pm$ 0.06 & 35.90 $\pm$ 0.04 & 27.52 $\pm$ 0.23 & 84.08 $\pm$ 0.10 & 3.58 $\pm$ 0.01 & 17.89 $\pm$ 0.02 \\
        & & no & 62.21 $\pm$ 0.04 & 61.79 $\pm$ 0.14 & N/A & N/A & 75.36 $\pm$ 0.15 & 68.88 $\pm$ 0.13 & N/A & N/A \\
    \midrule
        \multirow{3}{*}{Vicuna} & \multirow{3}{*}{13B} & original & 94.94 $\pm$ 0.19 & 95.47 $\pm$ 0.12 & \textbf{31.90 $\pm$ 0.16} & \textbf{46.84 $\pm$ 0.06} & 87.57 $\pm$ 0.30 & 79.89 $\pm$ 0.31 & 7.30 $\pm$ 0.07 & 22.67 $\pm$ 0.07 \\
        & & edited & 13.00 $\pm$ 0.11 & 91.05 $\pm$ 0.15 & 30.72 $\pm$ 0.18 & 45.71 $\pm$ 0.08 & 33.43 $\pm$ 0.43 & 78.87 $\pm$ 0.41 & 6.65 $\pm$ 0.04 & 22.07 $\pm$ 0.04 \\
        & & no & 59.32 $\pm$ 0.46 & 58.98 $\pm$ 0.61 & N/A & N/A & 69.43 $\pm$ 0.50 & 64.28 $\pm$ 0.17 & N/A & N/A \\
    \midrule
        \multirow{3}{*}{Baichuan2} & \multirow{3}{*}{13B} & original & 90.18 $\pm$ 0.19 & 89.82 $\pm$ 0.02 & 26.76 $\pm$ 0.00 & 42.20 $\pm$ 0.03 & 81.30 $\pm$ 0.28 & 72.50 $\pm$ 0.07 & 6.82 $\pm$ 0.02 & 21.80 $\pm$ 0.04 \\
        & & edited & 15.15 $\pm$ 0.13 & 86.88 $\pm$ 0.02 & 25.82 $\pm$ 0.05 & 41.24 $\pm$ 0.05 & 36.57 $\pm$ 0.43 & 71.26 $\pm$ 0.21 & 6.56 $\pm$ 0.03 & 21.50 $\pm$ 0.04 \\
        & & no & 62.89 $\pm$ 0.11 & 62.54 $\pm$ 0.19 & N/A & N/A & 64.72 $\pm$ 0.46 & 58.90 $\pm$ 0.07 & N/A & N/A \\
    \bottomrule
    \end{tabular}}
    \caption{Results of response quality evaluation. ``Original'', ``edited'', and ``no'' represent providing models with original right contexts, edited wrong contexts, and no contexts, respectively. The best result in each column is highlighted in \textbf{bold}.}
    \label{tab:quality_evaluation}
\end{table*}

\begin{table*}[t]
    \centering
    \resizebox{0.9\linewidth}{!}{
    \begin{tabular}{@{}lc|ccc|c|c|c@{}}
    \toprule
         \multirow{3}{*}{Models} & \multirow{3}{*}{Size} & \multicolumn{3}{c|}{EventKG} & \multicolumn{3}{c}{UJ} \\
         & & QA-A & QA-NA & Text Generation & QA-A & QA-NA & Text Generation \\
         & & M-Rate $\downarrow$ & M-Rate $\downarrow$ & R-Rate $\downarrow$ & M-Rate $\downarrow$ & M-Rate $\downarrow$ & R-Rate $\downarrow$\\
    \midrule
         ChatGLM2 & 6B & \textbf{80.87 $\pm$ 0.09} & 5.36 $\pm$ 0.11 & \textbf{85.09 $\pm$ 0.08} & \textbf{55.83 $\pm$ 0.61} & 11.25 $\pm$ 0.14 & 65.91 $\pm$ 0.25 \\
         Llama2 & 13B & 84.92 $\pm$ 0.12 & \textbf{3.60 $\pm$ 0.07} & 91.97 $\pm$ 0.09 & 68.63 $\pm$ 0.17 & \textbf{10.68} $\pm$ 0.30 & 74.06 $\pm$ 0.56 \\
         Vicuna & 13B & 84.97 $\pm$ 0.08 & 6.43 $\pm$ 0.27 & 90.94 $\pm$ 0.19 & 60.49 $\pm$ 0.05 & 12.28 $\pm$ 0.38 & 69.72 $\pm$ 0.72 \\
         Baichuan2 & 13B & 83.05 $\pm$ 0.24 & 9.16 $\pm$ 0.10 & 90.49 $\pm$ 0.14 & 57.19 $\pm$ 0.43 & 17.99 $\pm$ 0.06 & \textbf{55.19 $\pm$ 0.14} \\
    \bottomrule
    \end{tabular}}
    \caption{Results of mistakes detection evaluation. All experiments are conducted under the ``edited'' setting.  The best result in each column is highlighted in \textbf{bold}.}
    \label{tab:robustness_evaluation}
\end{table*}

\subsection{Metrics}
We assess the models' performance in two aspects: 
1) Can models still generate high-quality responses even if interfered with by counterfactual information? (Response Quality Aspect) 2) Can models resist the counterfactual information in the contexts? (Robustness Aspect)

The notations used in the following definitions of metrics are shown in Table~\ref{tab:notation}.

For question answering, we use accuracy and Misleading Rate (M-Rate) to evaluate models' performance in the two aspects, respectively. Misleading Rate is defined as:

\begin{small}
\begin{equation}
    \textrm{M-Rate}=\frac{\sum_{i=1}^{N_Q}\mathbb{I}(p_i^e\neq a_i^e \land p_i^n=a_i^n)}{\sum_{i=1}^N\mathbb{I}(p_i^n=a_i^n)}
\end{equation}
\end{small}

In other words, M-Rate is the proportion of the queries that the model answers wrongly with edited contexts in all queries that the model can answer correctly without external knowledge.

For text generation, we choose BLEU~\citep{papineni-etal-2002-bleu} and ROUGE-L~\citep{lin-2004-rouge} as the evaluation metrics for response quality. For the evaluation of robustness, we use Mistake Reappearance Rate (R-Rate) for EventKG which is defined as:

\begin{small}
\begin{equation}
    \textrm{R-Rate}=\frac{\sum_{i=1}^{N_T}\sum_{j=1}^{|\mathbb{E}_i^T|}\mathbb{I}(\mathbb{E}_{ij}^T\in\mathbb{A}_i^T \land \mathbb{O}_{ij}^T\in\mathbb{A}_i^T)}{\sum_{i=1}^{N_T}\sum_{j=1}^{|\mathbb{E}_i^T|}\mathbb{I}(\mathbb{O}_{ij}^T\in\mathbb{A}_i^T)}
\end{equation}
\end{small}

Mistake Reappearance Rate is the proportion of edited words appearing in the model's outputs in all edited words. When it comes to UJ, we adjust the definition of R-Rate because the original words and edited words in UJ cannot be matched exactly one by one. Thus R-Rate for UJ will be as:

\begin{small}
\begin{equation}
    \textrm{R-Rate}=\frac{\sum_{i=1}^{N_T}\sum_{j=1}^{|\mathbb{E}_i^T|}\mathbb{I}(\mathbb{E}_{ij}^T\in\mathbb{A}_i^T)/\sum_{i=1}^{N_T}|\mathbb{E}_i^T|}{\sum_{i=1}^{N_T}\sum_{j=1}^{|\mathbb{O}_i^T|}\mathbb{I}(\mathbb{O}_{ij}^T\in\mathbb{A}_i^T)/\sum_{i=1}^{N_T}|\mathbb{O}_i^T|}
\end{equation}
\end{small}

\subsection{Models}
We use the following large language models for our experimental evaluation.
\begin{itemize}

\item \textbf{ChatGLM2}~\citep{du-etal-2022-glm}: a series of AI chat models from THUDM. We use ChatGLM2-6B model in our experiments.

\item \textbf{Llama2}~\citep{touvron2023llama}: a large language model released by Meta AI, which supports text completion and chat completion. We use Llama2-13b-chat in specific.

\item \textbf{Vicuna}~\citep{vicuna2023}: a series of language models which is tuned from Llama models. We adopt the vicuna-13b-v1.5 model.

\item \textbf{Baichuan2}~\citep{yang2023baichuan}: the second generation of LLM from Baichuan Intelligence and supports both Chinese and English. We use the Baichuan2-13B-Chat version.

\end{itemize}
\subsection{Results of Response Quality Evaluation}
The results of evaluation metrics in response quality aspect are shown in Table \ref{tab:quality_evaluation}.

\paragraph{Question Answering} In general, adding counterfactual mistakes into contexts will lead to a drop in accuracy. In comparison, the accuracy drop in QA-A is much more significant than that in QA-NA. The accuracy with edited contexts in QA-A is even lower than that with no contexts provided. Though editing non-answer texts will not directly affect the answers, models still suffer a slight decrease in accuracy. Considering specific datasets, the influence of counterfactual mistakes is severer on EventKG than on UJ. A possible explanation is that the mistakes we add in UJ are mainly logical errors, which can contradict other information in the context. As a result, models will not be affected easily by these incoherent contexts and choose to believe in the intrinsic knowledge in themselves. In comparison, the counterfactual mistakes we add to EventKG are mainly on entities, which will not affect the consistency of the contexts. Thus there is a higher possibility for models to trust the seemingly coherent contexts.

\paragraph{Text Generation} Editing some words and phrases in the source texts does not severely influence the performance of models in BLEU and ROUGE-L because only a few words in the models' generations will change. For text generation, traditional metrics cannot truly reflect the harmfulness of the counterfactual information in the source texts. Therefore, our metrics in the robustness aspect are significant for truthful evaluation of models' robustness to counterfactual contexts.

\subsection{Results of Robustness Evaluation}
The results of evaluation metrics in the robustness aspect are shown in Table \ref{tab:robustness_evaluation}.

\paragraph{Question Answering} We can clearly see that directly editing answer texts has a much higher probability of misleading the models. It is interesting that editing answer/non-answer texts has totally different impacts on the two datasets. The M-Rate of QA-A in UJ is much lower than that in EventKG, which can be explained by the same reason in the previous section that models will show less trust in these errors in UJ due to inner-text contradictions. However, the M-Rate in UJ surpasses that in EventKG when it comes to QA-NA. When we edit the non-answer texts, the mistakes may contradict the answers in the contexts, which will render the models to disbelieve the actually right answers.

\paragraph{Text Generation} The high R-Rate indicates that counterfactual errors in the contexts will be probably adopted by the models and appear in the final outputs. On EventKG, this problem is extremely severe, where models fail to neglect over 85\% of the mistakes in all cases and present them to users in their responses. On UJ, R-Rate is comparably lower. Since we ask models to generate a definition for each scientific term in only one sentence and a large portion of words will not appear in the responses, we speculate that this phenomenon can be attributed to the limited number of words the models can adopt in their outputs.

\begin{table*}[t]
    \centering
    \resizebox{0.9\linewidth}{!}{
    \begin{tabular}{@{}lcl|cccc|cccc@{}}
    \toprule
         \multirow{3}{*}{Models} & \multirow{3}{*}{Size} & \multirow{3}{*}{Methods} & \multicolumn{4}{c|}{EventKG} & \multicolumn{4}{c}{UJ} \\
         & & & QA-A & QA-NA & \multicolumn{2}{c|}{Text Generation} & QA-A & QA-NA & \multicolumn{2}{c}{Text Generation} \\
         & & & ACC $\uparrow$ & ACC $\uparrow$ & BLEU $\uparrow$ & ROUGE-L $\uparrow$ & ACC $\uparrow$ & ACC $\uparrow$ & BLEU $\uparrow$ & ROUGE-L $\uparrow$ \\
    \midrule
        \multirow{3}{*}{ChatGLM2} & \multirow{3}{*}{6B} & Baseline & 15.71 $\pm$ 0.07 & 90.99 $\pm$ 0.03 & 29.07 $\pm$ 0.05 & 44.17 $\pm$ 0.01 & 36.05 $\pm$ 0.25 & 78.68 $\pm$ 0.10 & 8.14 $\pm$ 0.02 & 24.21 $\pm$ 0.02 \\
        & & Prompt & \textbf{21.04 $\pm$ 0.23} & 82.91 $\pm$ 0.13 & 28.79 $\pm$ 0.04 & 43.42 $\pm$ 0.04 & 35.18 $\pm$ 0.38 & 72.97 $\pm$ 0.20 & 7.89 $\pm$ 0.02 & 23.98 $\pm$ 0.02 \\
        & & DoLa & \textbf{15.77 $\pm$ 0.05} & \textbf{91.16 $\pm$ 0.04} & 28.98 $\pm$ 0.07 & \textbf{44.24 $\pm$ 0.05} & 35.90 $\pm$ 0.34 & \textbf{79.48 $\pm$ 0.13} & 8.10 $\pm$ 0.02 & \textbf{24.31 $\pm$ 0.02} \\
    \midrule
        \multirow{3}{*}{Llama2} & \multirow{3}{*}{13B} & Baseline & 12.80 $\pm$ 0.10 & 94.29 $\pm$ 0.06 & 18.31 $\pm$ 0.06 & 35.90 $\pm$ 0.04 & 27.52 $\pm$ 0.23 & 84.08 $\pm$ 0.10 & 3.58 $\pm$ 0.01 & 17.89 $\pm$ 0.02 \\
        & & Prompt & \textbf{15.37 $\pm$ 0.07} & 90.82 $\pm$ 0.18 & 15.04 $\pm$ 0.03 & 33.19 $\pm$ 0.04 & \textbf{30.33 $\pm$ 0.09} & 82.53 $\pm$ 0.02 & \textbf{3.93 $\pm$ 0.02} & \textbf{18.27 $\pm$ 0.03} \\
        & & DoLa & \textbf{12.86 $\pm$ 0.06} & 94.06 $\pm$ 0.07 & \textbf{18.54 $\pm$ 0.08} & \textbf{35.98 $\pm$ 0.04} & \textbf{27.63 $\pm$ 0.01} & 83.98 $\pm$ 0.03 & \textbf{3.87 $\pm$ 0.01} & \textbf{18.32 $\pm$ 0.01} \\
    \midrule
        \multirow{3}{*}{Vicuna} & \multirow{3}{*}{13B} & Baseline & 13.00 $\pm$ 0.11 & 91.05 $\pm$ 0.15 & 30.72 $\pm$ 0.18 & 45.71 $\pm$ 0.08 & 33.43 $\pm$ 0.43 & 78.87 $\pm$ 0.41 & 6.65 $\pm$ 0.04 & 22.07 $\pm$ 0.04 \\
        & & Prompt & \textbf{14.41 $\pm$ 0.11} & 88.10 $\pm$ 0.15 & \textbf{31.31 $\pm$ 0.16} & \textbf{45.85 $\pm$ 0.09} & \textbf{35.57 $\pm$ 0.45} & 78.23 $\pm$ 0.38 & \textbf{7.15 $\pm$ 0.05} & \textbf{22.86 $\pm$ 0.05} \\
        & & DoLa & 12.35 $\pm$ 0.14 & \textbf{92.08 $\pm$ 0.07} & \textbf{31.37 $\pm$ 0.10} & \textbf{45.85 $\pm$ 0.04} & \textbf{34.93 $\pm$ 0.18} & 78.27 $\pm$ 0.11 & 6.57 $\pm$ 0.06 & 22.05 $\pm$ 0.05 \\
    \midrule
        \multirow{3}{*}{Baichuan2} & \multirow{3}{*}{13B} & Baseline & 15.15 $\pm$ 0.13 & 86.88 $\pm$ 0.02 & 25.82 $\pm$ 0.05 & 41.24 $\pm$ 0.05 & 36.57 $\pm$ 0.43 & 71.26 $\pm$ 0.21 & 6.56 $\pm$ 0.03 & 21.50 $\pm$ 0.04 \\
        & & Prompt & \textbf{15.89 $\pm$ 0.14} & 84.54 $\pm$ 0.22 & \textbf{27.96 $\pm$ 0.05} & \textbf{42.32 $\pm$ 0.08} & \textbf{37.52 $\pm$ 0.25} & 67.81 $\pm$ 0.09 & \textbf{6.92 $\pm$ 0.04} & \textbf{22.22 $\pm$ 0.05} \\
        & & DoLa & \textbf{16.70 $\pm$ 0.19} & \textbf{87.39 $\pm$ 0.09} & \textbf{28.14 $\pm$ 0.12} & \textbf{43.39 $\pm$ 0.09} & \textbf{40.95 $\pm$ 0.39} & \textbf{75.61 $\pm$ 0.19} & \textbf{7.33 $\pm$ 0.03} & \textbf{22.86 $\pm$ 0.02} \\
    \midrule
        \multirow{3}{*}{Average} & \multirow{3}{*}{/} & Baseline & 14.17 & 90.80 & 25.98 & 41.76 & 33.39 & 78.22 & 6.23 & 21.42 \\
        & & Prompt & \textbf{16.67} & 86.59 & 25.78 & 41.20 & \textbf{34.65} & 75.39 & \textbf{6.47} & \textbf{21.83} \\
        & & DoLa & \textbf{14.42} & \textbf{91.17} & \textbf{26.76} & \textbf{42.36} & \textbf{34.85} & \textbf{79.33} & \textbf{6.47} & \textbf{21.88} \\
    \bottomrule
    \end{tabular}}
    \caption{Results of the prompt and DoLa methods for improving response quality. Better results compared to the corresponding baseline results are highlighted in \textbf{bold}.}
    \label{tab:improvement_aspect1}
\end{table*}

\begin{table*}[t]
    \centering
    \resizebox{0.9\linewidth}{!}{
    \begin{tabular}{@{}lcl|ccc|ccc@{}}
    \toprule
         \multirow{3}{*}{Models} & \multirow{3}{*}{Size} & \multirow{3}{*}{Methods} & \multicolumn{3}{c|}{EventKG} & \multicolumn{3}{c}{UJ} \\
         & & & QA-A & QA-NA & Text Generation & QA-A & QA-NA & Text Generation \\
         & & & M-Rate $\downarrow$ & M-Rate $\downarrow$ & R-Rate $\downarrow$ & M-Rate $\downarrow$ & M-Rate $\downarrow$ & R-Rate $\downarrow$\\
    \midrule
        \multirow{3}{*}{ChatGLM2} & \multirow{3}{*}{6B} & Baseline & 80.87 $\pm$ 0.09 & \phantom{0}5.36 $\pm$ 0.11 & 85.09 $\pm$ 0.08 & 55.83 $\pm$ 0.61 & 11.25 $\pm$ 0.14 & 65.91 $\pm$ 0.25 \\
        & & Prompt & \textbf{74.90 $\pm$ 0.27} & 11.49 $\pm$ 0.09 & \textbf{82.83 $\pm$ 0.53} & 56.60 $\pm$ 0.22 & 16.39 $\pm$ 0.24 & \textbf{65.61 $\pm$ 0.39} \\
        & & DoLa & \textbf{80.74 $\pm$ 0.06} & \textbf{\phantom{0}4.91 $\pm$ 0.06} & 85.16 $\pm$ 0.41 & 55.88 $\pm$ 0.50 & \textbf{10.91 $\pm$ 0.14} & \textbf{65.47 $\pm$ 0.25} \\
    \midrule
        \multirow{3}{*}{Llama2} & \multirow{3}{*}{13B} & Baseline & 84.92 $\pm$ 0.12 & \phantom{0}3.60 $\pm$ 0.07 & 91.97 $\pm$ 0.09 & 68.63 $\pm$ 0.17 & 10.68 $\pm$ 0.30 & 74.06 $\pm$ 0.56 \\
        & & Prompt & \textbf{81.41 $\pm$ 0.20} & \phantom{0}6.14 $\pm$ 0.07 & \textbf{90.87 $\pm$ 0.19} & \textbf{66.00 $\pm$ 0.15} & 11.56 $\pm$ 0.12 & \textbf{69.12 $\pm$ 0.04} \\
        & & DoLa & \textbf{84.85 $\pm$ 0.06} & \phantom{0}3.74 $\pm$ 0.10 & 92.03 $\pm$ 0.08 & \textbf{68.55 $\pm$ 0.19} & 10.71 $\pm$ 0.27 & 75.86 $\pm$ 0.18 \\
    \midrule
        \multirow{3}{*}{Vicuna} & \multirow{3}{*}{13B} & Baseline & 84.97 $\pm$ 0.08 & \phantom{0}6.43 $\pm$ 0.27 & 90.94 $\pm$ 0.19 & 60.49 $\pm$ 0.05 & 12.28 $\pm$ 0.38 & 69.72 $\pm$ 0.72 \\
        & & Prompt & \textbf{82.71 $\pm$ 0.23} & \phantom{0}8.76 $\pm$ 0.26 & \textbf{89.84 $\pm$ 0.07} & \textbf{58.33 $\pm$ 0.26} & 12.88 $\pm$ 0.21 & \textbf{58.14 $\pm$ 0.32} \\
        & & DoLa & 85.98 $\pm$ 0.09 & \textbf{\phantom{0}5.98 $\pm$ 0.17} & \textbf{90.93 $\pm$ 0.08} & \textbf{58.99 $\pm$ 0.29} & 12.90 $\pm$ 0.17 & 72.16 $\pm$ 0.23 \\
    \midrule
        \multirow{3}{*}{Baichuan2} & \multirow{3}{*}{13B} & Baseline & 83.05 $\pm$ 0.24 & \phantom{0}9.16 $\pm$ 0.10 & 90.49 $\pm$ 0.14 & 57.19 $\pm$ 0.43 & 17.99 $\pm$ 0.06 & 55.19 $\pm$ 0.14 \\
        & & Prompt & \textbf{81.56 $\pm$ 0.11} & 10.61 $\pm$ 0.28 & 90.71 $\pm$ 0.18 & \textbf{55.71 $\pm$ 0.41} & 21.56 $\pm$ 0.07 & 55.43 $\pm$ 0.32 \\
        & & DoLa & \textbf{82.53 $\pm$ 0.09} & 10.03 $\pm$ 0.12 & 91.74 $\pm$ 0.14 & \textbf{53.40 $\pm$ 0.23} & \textbf{15.13 $\pm$ 0.22} & 63.89 $\pm$ 0.41 \\
    \midrule
        \multirow{3}{*}{Average} & \multirow{3}{*}{/} & Baseline & 83.45 & 6.14 & 89.62 & 60.54 & 13.05 & 66.22 \\
        & & Prompt & \textbf{80.15} & 9.25 & \textbf{88.56} & \textbf{59.16} & 15.60 & \textbf{62.08} \\
        & & DoLa & 83.52 & 6.16 & 89.97 & \textbf{59.20} & \textbf{12.41} & 69.35 \\
    \bottomrule
    \end{tabular}}
    \caption{Results of the prompt and DoLa methods for enhancing models' robustness to counterfactual contexts. Better results compared to the corresponding baseline results are highlighted in \textbf{bold}.}
    \label{tab:improvement_aspect2}
\end{table*}

\section{Measures to Enhance Robustness}
\subsection{Methods}
In this section, we will adopt several methods in order to enhance models' robustness when counterfactual information in the contexts contradicts the models' own knowledge. Concretely, we select two different kinds of methods and evaluate their performances with edited contexts given in the same settings as those in \S~\ref{sec:eval}.

\paragraph{Prompting} It is a simple but intuitive way that we explicitly ask the model to neglect counterfactual mistakes in the queries. Specifically, we add an instruction at the end of each query, which asks the models to believe in themselves when the external information contradicts their own knowledge.

\paragraph{Inference Intervention} To mitigate hallucination in LLMs, recent studies intervene in models' inference processes to enhance generation quality, such as ITI~\citep{Li2023InferenceTimeIE}, DoLa~\citep{Chuang2023DoLaDB}, representation engineering~\citep{zou2023representation}, and activation addition~\citep{turner2023activation}. We test DoLa in our experiments as the representative of this type of methods.

\subsection{Results}
The evaluation results of the two methods are shown in Table \ref{tab:improvement_aspect1} and Table \ref{tab:improvement_aspect2}. 

\paragraph{Prompting} In QA-A, explicitly demanding models to neglect counterfactual information can steadily benefit models' response quality and robustness, which can be proven by the rise in accuracy and the drop in the M-Rate. In text generation, this method does not obviously increase BLEU and ROUGE-L metrics but results in a significant improvement in R-Rate. However, the prompting method harms both models' response quality and robustness to counterfactual mistakes at the same time in QA-NA. Considering that it is a straightforward and cost-efficient method to adjust the prompt, this method still has its value in some cases.

\paragraph{Inference Intervention} In the aspect of response quality, DoLa successfully improves models' answers in all tasks. It proves that inference intervention methods can actually help to generate more accurate answers. However, DoLa fails to enhance the models' robustness, which is indicated by the moving trend of R-Rate in the robustness aspect. This phenomenon suggests that DoLa cannot deal with the contradiction between external texts and internal knowledge and is not suitable for the problem we focus on.

In general, neither of the methods can bring steady and significant improvements to models' response quality and robustness at the same time. There is still a high possibility that models will be misled by those counterfactual mistakes existing in contexts from external knowledge bases or the Internet and finally generate wrong answers for user queries. The results also prove that the problem we identify in this paper cannot be solved by existing methods and deserves further studies in the future.

\begin{figure}[t]
    \centering
    \subfloat[EventKG]{
        \includegraphics[width=0.9\linewidth]{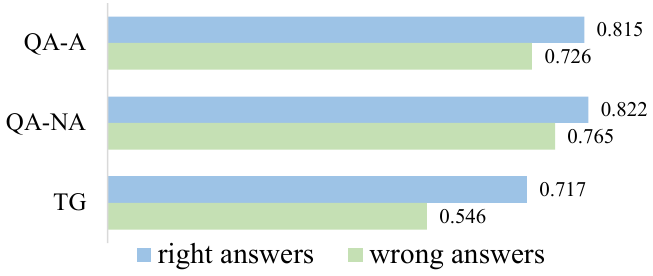}
    }\\
    \subfloat[UJ]{
        \includegraphics[width=0.9\linewidth]{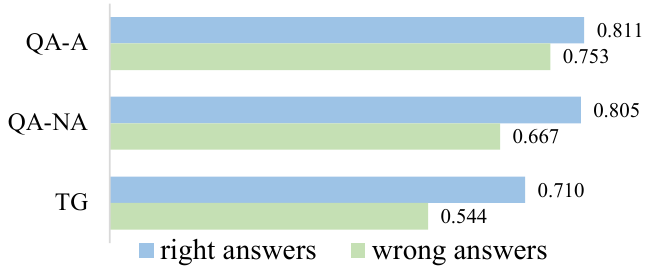}
    }
    \caption{The confidence of ChatGLM2 on its right and wrong answers, respectively.}
    \label{fig:confidence}
\end{figure}

\section{Analyses} 
\subsection{Observation on Model Confidence}
\label{sec:confidence}
To study the influence of counterfactual contexts on models' final outputs quantitatively, we compare the average generation probabilities of all tokens in wrong answers and right answers generated by ChatGLM2, respectively. For QA-A and QA-NA, we only consider the samples that the model can answer correctly with no contexts provided but will give wrong predictions when given counterfactual information. For text generation, we compare the probabilities of original words/phrases and those of edited words/phrases.

The results are shown in Figure~\ref{fig:confidence}. We observe that the model's confidence in its answers will significantly drop with the interference of counterfactual contexts, especially for text generation. This phenomenon suggests that it is plausible to guide models to generate accurate responses according to the model's confidence in future research.

\begin{table}[t]
    \centering\resizebox{1.0\linewidth}{!}{
    \begin{tabular}{l|l}
    \toprule
        Context & \makecell[l]{The Trial of Joan of Arc took place during the 15th century\\ and was a legal proceeding against Joan of Arc. Joan of Arc\\ was a French military leader who served under Charles VII\\ during the Hundred Years' War. The trial began on January 9,\\ 1431, and lasted until May 29, 1431. (\textcolor{red}{The trial began on}\\ \textcolor{red}{January 10, 1432, and lasted until May 30, 1432}) Joan of Arc\\ was the defendant in the trial, while \textcolor{blue}{Pierre Cauchon} acted as\\ both the prosecutor and the judge. She was charged with heresy.\\} \\
    \midrule
        Question & Who was the judge in the Trial of Joan of Arc? \\
        Options & 1) Pierre Cauchon; 2) Jean d'Estivet; \\
        Original Answer & 1) Pierre Cauchon; \\
        Prompt Answer & 2) Jean d'Estivet; \\
    \bottomrule
    \end{tabular}}
    \caption{A case where ChatGLM2 gives a wrong answer with edited contexts provided. The sentence in \textcolor{red}{red} is the edited one with time information changed. The answer to the question appearing in the context is in \textcolor{blue}{blue}.}
    \label{tab:case}
\end{table}

\subsection{Case Study}
We can observe from Table~\ref{tab:improvement_aspect1} and Table~\ref{tab:improvement_aspect2} that the prompt method achieves an improvement in QA-A but causes a drop in performance in QA-NA. A typical case, in which the model answers the question incorrectly after adding a new sentence into the prompt while it could originally give a right prediction, is shown in Table~\ref{tab:case}.
In this case, we edit the beginning and end time of the trial, which will not affect the answer to the question about the judge of the trial. After we add a sentence that instructs the model to neglect counterfactual information in the context, the model changes its answer from the right one \textit{Pierre Cauchon}, which has been clearly stated in the context, to another name that even does not appear in the context.

We speculate that the instruction we add to the prompt will reduce models' trust in the external knowledge. In QA-A, the models will thus disbelieve the edited wrong answer in the context. However, in QA-NA, models' confidence in the right answer appearing in the context will also fall due to the instruction. This case indicates that it is challenging to solve this problem by just modifying the prompt.

\section{Related Works}

\paragraph{Hallucination in LLMs}
Although LLMs excel at generating fluent natural language, studies show that they are subject to the problem of hallucination, which means that texts generated by the models often contain information that is irrelevant to user inputs, conflicting with previous responses, or unfaithful to established world knowledge~\citep{Ji2022SurveyOH,rawte2023survey,Zhang2023SirensSI,Huang2023ASO}. 
Some studies aim to mitigate the issue of hallucination by incorporating additional information into the generation procedure, such as Web corpora~\citep{shuster-etal-2021-retrieval-augmentation,Huo2023RetrievingSE,Yu2023ImprovingLM}, knowledge graphs~\citep{Ji2022RHOR}, and external tools~\citep{Gou2023CRITICLL}.
Another line of work focuses on improving the decoding strategy of LLMs, such as careful prompt design~\citep{Mndler2023SelfcontradictoryHO}, sampling multiple responses~\citep{Manakul2023SelfCheckGPTZB}, and manipulating internal model states~\citep{Chuang2023DoLaDB,Li2023InferenceTimeIE,Azaria2023TheIS,zou2023representation,turner2023activation}. 
Efforts have also been made to establish benchmarks for comprehensively evaluating the truthfulness and coherence of language models~\citep{Liu2021ATR,Lin2021TruthfulQAMH,Liang2023HolisticEO,Li2023HaluEvalAL}.

\paragraph{Sycophancy in LLMs}
Sycophancy refers to the tendency of LLMs to tailor their responses in order to seek human approval~\citep{perez-etal-2023-discovering,Wei2023SimpleSD}. 
For example, they often change their answers when their responses are questioned or cater to specific political views of the user. 
Previous work~\citep{sharma2023understanding} attributes sycophantic behavior to the use of preference models for LLM alignment during the pre-training stage. 
In contrast to existing work on sycophancy, we investigate the specific problem of LLM robustness against misinformation in user inputs and curate a benchmark for its systematic evaluation.

\section{Conclusion}
In this paper, we focus on a new problem that the contradiction between extrinsic information and intrinsic knowledge will mislead LLMs and result in low-quality responses.
Due to the lack of suitable benchmarks and evaluation metrics, we construct a benchmark RECALL and design two task-specific metrics to evaluate the models' robustness. The evaluation results indicate that current LLMs are vulnerable to misinformation in the contexts related to user queries and will be easily misled when the mistakes in the contexts contradict their own knowledge. We further attempt to improve models' performance with two different kinds of methods but achieve unsatisfactory results, which suggests that existing approaches fail to solve the problem we identify and further studies deserve to be conducted to find more effective methods.
In conclusion, the benchmark we construct in this paper provides a unified and reliable standard for the evaluation of LLMs' robustness against counterfactual information, and the two new metrics we propose will become trustworthy criteria for future methods proposed to solve this problem.

\section*{Acknowledgements}
We sincerely thank Deli Chen for his valuable suggestions. Xu Sun is the corresponding author.

\bibliography{anthology,custom}

\end{document}